\documentclass[lettersize,journal]{IEEEtran}
\usepackage{amsmath,amsfonts}
\usepackage{algorithmic}
\usepackage{algorithm}
\usepackage[colorlinks,linkcolor=red,anchorcolor=blue,citecolor=blue]{hyperref}
\usepackage{bbm}
\usepackage{amssymb}
\usepackage{multirow}
\usepackage{array}
\usepackage[caption=false,font=normalsize,labelfont=sf,textfont=sf]{subfig}
\usepackage{textcomp}
\usepackage{stfloats}
\usepackage{url}
\usepackage{verbatim}
\usepackage{graphicx}
\usepackage{float}
\usepackage{cite}
\usepackage{listings}
\usepackage[dvipsnames]{xcolor}
\def\eg{{\it e.g.}}
\def\ie{{\it i.e.}}
\begin{document}

\title{Learning to Discover Forgery Cues for Face Forgery Detection}

\author{Jiahe Tian, Peng Chen, Cai Yu, Xiaomeng Fu, Xi Wang, Jiao Dai, Jizhong Han \thanks{Jiahe Tian, Cai Yu and Xiaomeng Fu are with the Institute of Information Engineering, Chinese Academy of Sciences, Beijing, China, and also with the School of Cyber Security, University of Chinese Academy of Sciences, Beijing, China (e-mail: tianjiahe@iie.ac.cn, caiyu@iie.ac.cn, fuxiaomeng@iie.ac.cn).
Peng Chen is with the RealAI Inc, Beijing, China(e-mail: peng.chen@realai.ai)
Xi Wang, Jiao Dai and Jizhong Han are with the Institute of Information Engineering, Chinese Academy of Sciences, Beijing, China(e-mail: wangxiboss@163.com,  daijiao@iie.ac.cn, hanjizhong@iie.ac.cn).

}}

\IEEEpubid{0000--0000/00\$00.00~\copyright~2021 IEEE}

\maketitle

\begin{abstract}

Locating manipulation maps, \ie, pixel-level annotation of forgery cues, is crucial for providing interpretable detection results in face forgery detection.
Related learning objects have also been widely adopted as auxiliary tasks to improve the classification performance of detectors whereas they require comparisons between paired real and forged faces to obtain manipulation maps as supervision.
This requirement restricts their applicability to unpaired faces and contradicts real-world scenarios.
Moreover, the used comparison methods annotate all changed pixels, including noise introduced by compression and upsampling.
Using such maps as supervision hinders the learning of exploitable cues and makes models prone to overfitting.
To address these issues, we introduce a weakly supervised model in this paper, named Forgery Cue Discovery (FoCus), to locate forgery cues in unpaired faces.
Unlike some detectors that claim to locate forged regions in attention maps, FoCus is designed to sidestep their shortcomings of capturing partial and inaccurate forgery cues.
Specifically, we propose a classification attentive regions proposal module to locate forgery cues during classification and a complementary learning module to facilitate the learning of richer cues.
The produced manipulation maps can serve as better supervision to enhance face forgery detectors.
Visualization of the manipulation maps of the proposed FoCus exhibits superior interpretability and robustness compared to  existing methods.
Experiments on five datasets and four multi-task models demonstrate the effectiveness of FoCus in both in-dataset and cross-dataset evaluations.
\end{abstract}

\begin{IEEEkeywords}
Multimedia forensics, Face forgery detection, Weakly supervised learning.
\end{IEEEkeywords}

\section{Introduction}
\IEEEPARstart{W}{ith} the remarkable progress of digital face-generation technology, it is becoming increasingly easy to
produce realistic fake faces\cite{perov2020deepfacelab,nirkin2019fsgan,Dual-agent}. 
Unluckily, digital face-generation technology also enables attackers to make face forgery media for malicious purposes.
The spread of face forgery media has aroused broad public concerns due to its potential negative impact.
Under this background, the need for effective and robust face forgery detection methods has become more urgent.

\begin{figure}[t]
\includegraphics[width=0.9\linewidth]{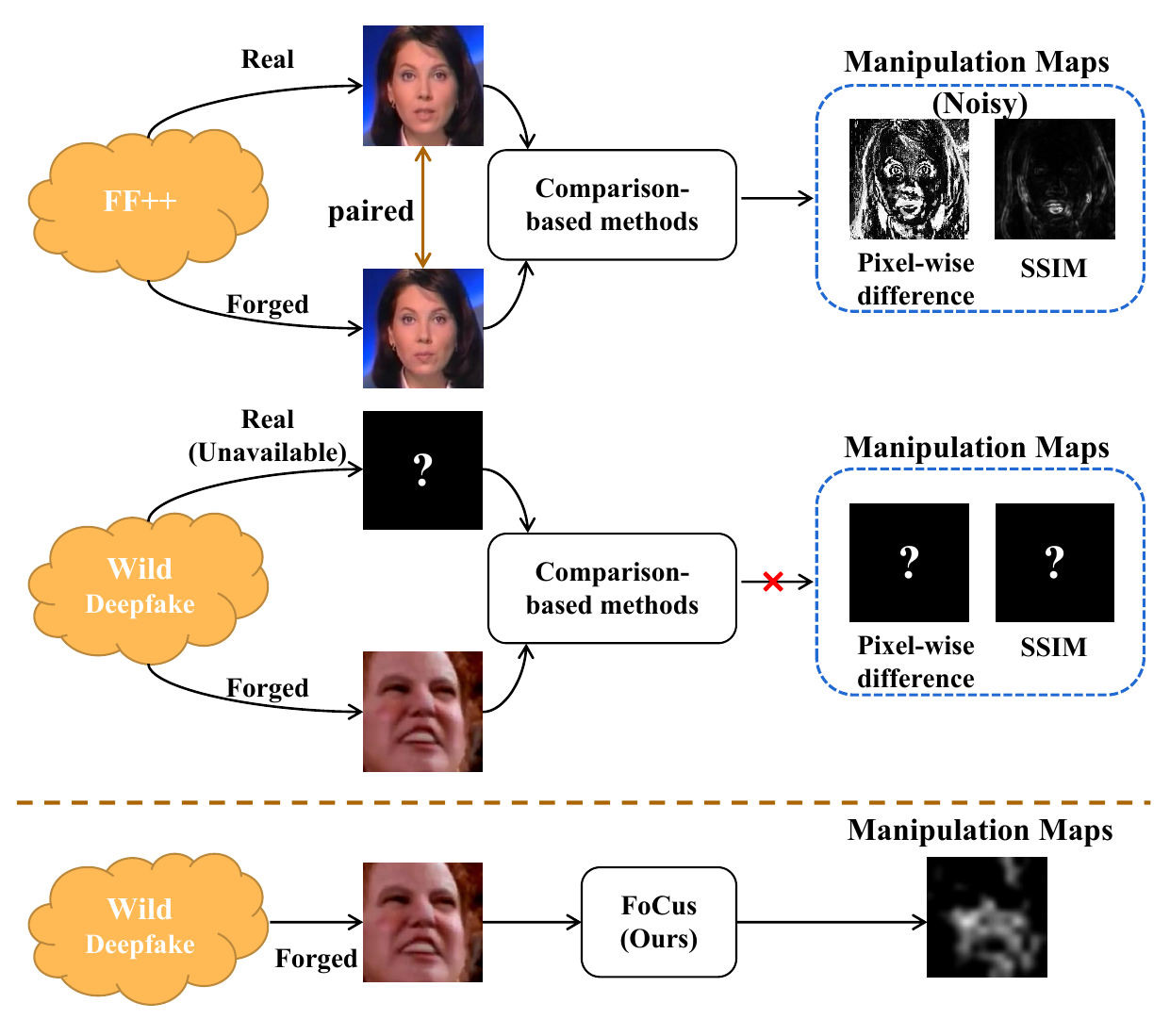}
\caption{Existing manipulation map generation methods rely on comparing paired faces. The generated maps are often noisy when treating globally disturbed images, leading to poor interpretability. We propose FoCus to generate manipulation maps in a weakly supervised manner.} 
\label{fig1}
\end{figure}

Face forgery detection is typically formulated as a binary classification problem that discriminates forged faces from real faces.
Recent works in this field primarily train deep neural networks equipped with specialized modules such as frequency forensics branch \cite{zhou2017two}, texture enhancing module \cite{gu2021exploiting}, and patch consistency learning module \cite{uiavit}.
However, these models inadvertently learn biased patterns while enhancing accuracy on the training set due to the subtle nature of forgery cues.
To mitigate this, some studies use auxiliary tasks to assist in the model's learning process and succeed in improving the classification performance.
Specifically, learning objects such as manipulated region prediction \cite{lttd,huang2020fakelocator, nguyen2019multi} and patch-level metric learning \cite{chen2021local, zhao2021learning} have been utilized to guide detection models to locate forged regions.
Namely, they conduct classification and forged region localization jointly in a multi-task learning manner.
To achieve this, pixel-level annotations are required to tag pixels or patches as real or fake. 
These annotations are typically obtained by comparing a forged face with its corresponding real face, as illustrated in Fig. \ref{fig1}.
Previous works commonly used comparison-based methods, \ie, Structural Similarity (SSIM) \cite{wang2004image} and pixel-wise difference, to generate manipulation maps. 
\IEEEpubidadjcol

While such tasks help to improve performance, their application scenarios are narrow as comparison-based methods require paired fake and real faces.
Consequently, multi-task models which require pixel-level annotations can only be trained on datasets with paired faces such as FaceForensics++\cite{rossler2019faceforensics++} and CelebDeepfake\cite{li2020celeb}.  
\IEEEpubidadjcol
Besides, these models amplify the gap between real-world inference scenarios and the training phase, making it impossible to expand their training data to vast forgeries on the internet \cite{ffiw, zi2020wilddeepfake} or train these methods to detect faces generated by the latest source-free synthesis techniques, \eg, Midjourney \cite{midjourney} and StableDiffusion \cite{rombach2021highresolution}.

In addition to the limited application scenarios, comparison-based methods are also sub-optimal in exploitability. 
Although they locate changed pixels, there is an inherent discrepancy between changed pixels and exploitable forgery cues.
A complete digital face-generation process involves post-processing such as compression and media encoding, which alters most pixels.
Besides, commonly used structures in face forgery algorithms, such as autoencoders and generative adversarial networks, have upsampling blocks, which introduce noise to the produced images. 
These noise patterns often fail to serve as general forgery cues. 
Thus, simply marking changed pixels is imprudent as it guides models to learn biased patterns instead of exploitable forgery cues. 
As shown in the upper part of Fig. \ref{fig1}, manipulation maps generated by comparison include pixels outside the face, \eg, background and clothing. 
Meanwhile, these maps lack interpretability and fail to assist humans in identifying forgeries.
Hence, we aim to generate interpretable and exploitable manipulation maps for unpaired faces.

Motivated by the need to generate manipulation maps for unpaired face forgeries, we introduce a weakly supervised model, named Forgery Cue Discovery (FoCus), to locate forgery cues in unpaired faces with binary classification labels.
We note that several detectors claimed their attention maps can locate forged regions\cite{uiavit,zhao2021multi,luo2021generalizing}.
However, their visualization results either simply cover the entire face or focus on one small area, which is far from accurate.
We argue this is due to the weakness of the attention map for omitting relatively weak forgery patterns in face forgeries.
Besides, when imposing constraints on attention maps, they tend to collapse to large areas to maintain the classification performance, making it difficult to expose forgery cues. 
To sidestep the weakness of generating manipulation maps through attention maps, we propose to locate classification attentive regions as forgery cues.
Specifically, a Classification Attentive Regions Proposal (CARP) module composed of a fully convolutional block and a classification attention map proposal operator are employed for locating forged regions.
To avoid capturing partial cues, we use max-pooling on multiple layers of feature maps to extract features for classification, which helps to expose multiple forgery regions.
Additionally, we introduce a Sobel branch to expose edge-related features following the practice in image forensics\cite{chen2021image}, which are robust to several corruptions.
To fuse located cues in the RGB and Sobel branches, we propose a Complementary Learning module to mine the complementary relationship between forgery cues in these two branches.
This module provides comprehensive patterns and prevents the locating process from collapsing to a single modality.
Finally, we generate manipulation maps based on the complementary relationship, which can serve as pixel-level annotations for aforementioned auxiliary tasks and provide interpretable forgery cues.

Since none of the current face forgery datasets offers ground truth for evaluating the generated maps, we evaluate these maps by using these maps as supervision in training multi-task detection models.
The used multi-task models vary in their architecture and design motivation.
The learning tasks for these models are face forgery classification and forged region localization.
We use different manipulation maps as supervision for the forged region localization task and compare the classification performance of these multi-task models. 
The quantitative experiments verify that the proposed FoCus exhibits better effectiveness compared to existing manipulation map generation methods in supervising multi-task face forgery detection models.  
The visualizations demonstrate the proposed FoCus locates interpretable forgery cues.

Our contributions can be summarized as follows:

$\bullet$ We introduce the problem of generating manipulation maps for face forgeries without using paired faces and a novel FoCus model for this problem. This allows us to broaden the training data of multi-task face forgery detection to vast unpaired forgeries from the internet.

$\bullet$ We carefully design two modules in FoCus to locate and fuse classification attention regions.
Compared to detectors which implicitly locate forgery cues through attention maps, our FoCus focuses on locating exploitable cues, making it more effective in supervising forged region localization tasks.

$\bullet$ We perform comprehensive experiments and visualizations to demonstrate the feasibility of generating manipulation maps through the proposed FoCus. The results validate the superior performance of FoCus  compared to previous methods in terms of generalization ability, interpretability, and robustness.

\section{Related Work}
\subsection{Face Forgery Detection}
A number of face forgery detection approaches have been proposed to mitigate the threats posed by the abuse of digital face generation technology. 
Early works in face forgery detection leverage abnormal biological artifacts to detect forgery \cite{li2018ictu,ciftci2020fakecatcher,yang2019exposing}, while more recent methods treat face forgery detection as a binary classification task and design various network structures for better performance\cite{afchar2018mesonet,zhao2021multi}.
For example, \cite{afchar2018mesonet} builds a shallow neural network to learn mesoscopic features that contain rich forgery cues, while
\cite{zhao2021multi} treats the feature maps after zero-mean normalization as textures and uses attention to fuse RGB and texture features. 
To enhance detection performance, another desirable strategy is introducing another image modality, including Steganalysis Rich Model (SRM) noise \cite{fridrich2012rich} in \cite{zhang2018face} and Discrete Cosine Transform (DCT) frequency \cite{ahmed1974discrete} in \cite{qian2020thinking}.
This scheme is also popular in other vision task domains, such as using Sobel-filtered images in image forensics  \cite{chen2021image} and using DCT coefficients in palmprint recognition\cite{LengLKB17}.
By analyzing the image from a different view, subtle forgery artifacts that are invisible in RGB images can be revealed. 
Inspired by these works, we use the Sobel\cite{sobel1968} filter to expose edge-related artifacts in face forgeries.

Several works in face forgery detection adopt multi-task learning with auxiliary supervision to improve classification performance. 
For instance, \cite{nguyen2019multi} incorporates a manipulation map prediction branch to guide the model to locate forged areas. 
\cite{chen2021local} and \cite{zhao2021learning} conduct metric learning on patch embeddings, which implicitly locate forged patches based on pixel-level annotation for forged regions.
To obtain pixel-level annotation for the manipulated regions, the aforementioned methods compare a forged face with the original face using pixel-wise differences or SSIM. 
However, the manipulation maps can be inaccessible as no original video is given, or the face positions, head poses and frame rate are inconsistent in paired videos.
In contrast, our proposed FoCus generates manipulation maps without relying on comparisons, which can be used as supervision for multi-task face forgery detectors.

\subsection{Weakly Supervised Semantic Segmentation}

Although most face forgery detection methods crop and resize the face images, making every face aligned, forgery cues can still locate in distinctive regions. 
We resort to weakly supervised semantic segmentation (WSSS) to generate manipulation maps using only class labels.
Semantic segmentation is an important task in computer vision that aims to segment objects of interest in an image\cite{lin2014microsoft,hoiem2009pascal,cordts2016cityscapes}.
Unlike supervised semantic segmentation, WSSS trains the network with weak labels, such as class labels \cite{pathak2015constrained,choe2020attention,hou2018self} or bounding boxes \cite{dai2015boxsup, khoreva2017simple}. 
Most WSSS models supervised by class labels generate and refine the class activation map (CAM) to approximate the segmentation map. 
Many of them follow the three-principle process introduced by \cite{kolesnikov2016seed}: \emph{seed}, \emph{expand}, and \emph{constrain}. 
Specifically, the seed attentive region is located by a classification model, and different schemes are adopted to expand and constrain the seed region to the semantic area.

\begin{figure*}[tp] 
\centering
{\includegraphics[width=\textwidth]{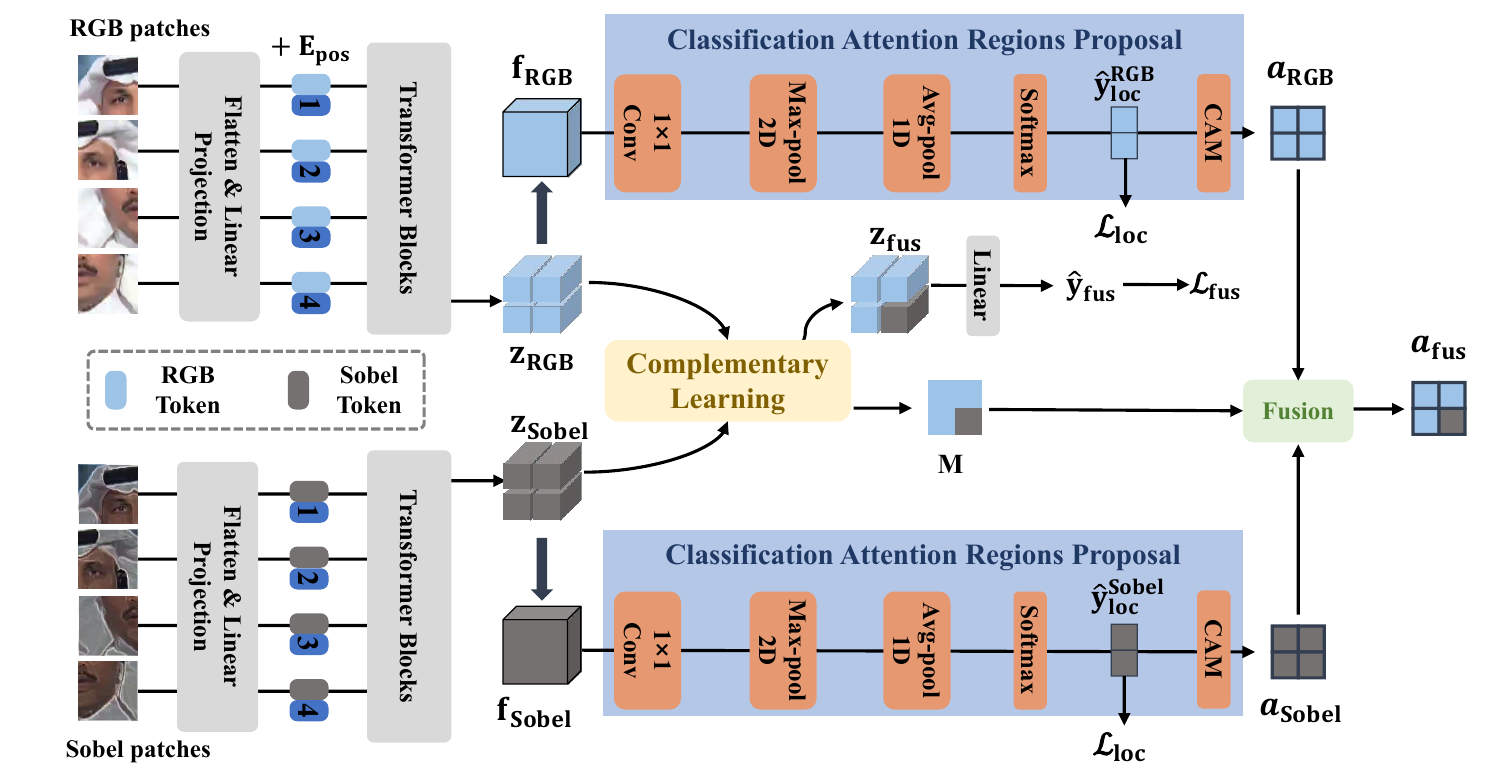}
\caption{The pipeline of the proposed FoCus. 
We use ViT as the backbone to encode RGB and Sobel inputs to $\mathbf{z}_{\rm RGB}$ and $\mathbf{z}_{\rm Sobel}$.  
The Classification Attentive Regions Proposal module is devised to locate forgery cues in both modalities to $\boldsymbol{a}_{\rm RGB}$ and $\boldsymbol{a}_{\rm Sobel}$.
The Complementary Learning module is devised to mine complementary nature between $\mathbf{z}_{\rm RGB}$ and $\mathbf{z}_{\rm Sobel}$, and then output a complementary mask $\mathbf{M}$ to fuse $\boldsymbol{a}_{\rm RGB}$ and $\boldsymbol{a}_{\rm Sobel}$ to $\boldsymbol{a}_{\rm fus}$ with Equation \ref{fusion_eqn}.
 $\boldsymbol{a}_{\rm fus}$ can serve as pixel-wise annotation for exploitable forgery cues. 
}
\label{fig2}
}
\end{figure*}
\section{Method}
This section details the proposed Forgery Cue Discovery (FoCus) model for manipulation maps generation where only image-level labels are available. 
We first give a brief preliminary for the Vision Transformer backbone for the proposed FoCus.
Then, we motivate and propose the Classification Attentive Regions Proposal (CARP) module for locating classification attentive region (CAR).
Finally, we describe the scheme for expanding the CAR maps with a complementary modality with the proposed Complementary Learning module.
The proposed FoCus is illustrated in Fig. \ref{fig2}.
\subsection{Vision Transformer}
Our model is built based on a Vision Transformer (ViT), which has become one of the most popular backbones in the field of computer vision.
Based on a transformer architecture, ViT models an image through the self-attention mechanism, achieving a global perception of the image.
We provide a preliminary for ViT here and refer to \cite{dosovitskiy2020image} for detailed information.

A vanilla ViT basically consists of a patch embedding layer and a sequence of transformer encoders.
In the patch embedding layer, an input image $\mathbf{x} \in R^{H\times W\times 3}$ is first split and flattened into patches $\mathbf{x}_p \in R^{N\times P^2\times D}$, where $(H, W)$ represents the height and width, $(P, P)$ is the patch size, $N =HW/P^2$ is the number of patches, and $D$ denotes patch embedding dimension. 
A learnable embedding $\mathbf{x}_{\text {class}}$ is further prepended to the sequence of patches to serve as the global representation.
To retain positional information, position encodings $\mathbf{E}_{\rm pos}$ are added to the sequence of patch embeddings:
\begin{equation}
\begin{aligned}
\mathbf{z}_0 & =\left[\mathbf{x}_{\text {class }} ; \mathbf{x}_p^1 \mathbf{E} ; \mathbf{x}_p^2 \mathbf{E} ; \cdots ; \mathbf{x}_p^N \mathbf{E}\right]+\mathbf{E}_{\rm pos}, 
\end{aligned}
\end{equation}
\noindent where $\mathbf{E} \in R^{\left(P^2 \cdot C\right) \times D}$ is the layer used to embed the flattened patches to tokens.
Each encoder consists of a multi-head self-attention layer (MHSA) and a multi-layer perceptron (MLP) block. 
In MHSA, the input tokens are projected to $\mathbf{q,k,v}$ as conducted in standard self-attention\cite{transformer}.
For each token in an input sequence, a weighted sum over all values $\mathbf{v}$ in the sequence is computed by:
\begin{equation}
\begin{aligned}
\centering
{[\mathbf{q}, \mathbf{k}, \mathbf{v}] }  &=\mathbf{z}_{\ell} \mathbf{U}_{q k v} \\
A  &=\operatorname{softmax}\left(\mathbf{q} \mathbf{k}^{\top} / \sqrt{D}\right) \\
\mathbf{z}_{\ell}^{\prime}  &=A \mathbf{v}
\end{aligned}
\end{equation}

\noindent where $\mathbf{U}_{q k v}$ is the projection layer, the attention weights $A$ are based on the pairwise similarity between $\mathbf{q}$ and $\mathbf{k}$, and $\mathbf{z}_{\ell}^{\prime}$ denotes the output of $l$-th MHSA.
After MHSA, $\mathbf{z}_{\ell}^{\prime}$ is sent to a LayerNorm (LN) layer and an MLP block. A residual connection is added after every block. 
The process can be formulated as: 
\begin{equation}
\begin{aligned}
\mathbf{z}_{\ell}^{\prime} & =\operatorname{MHSA}\left(\mathrm{LN}\left(\mathbf{z}_{\ell-1}\right)\right)+\mathbf{z}_{\ell-1}, & & \ell=1 \ldots L \\
\mathbf{z}_{\ell} & =\operatorname{MLP}\left(\mathrm{LN}\left(\mathbf{z}_{\ell}^{\prime}\right)\right)+\mathbf{z}_{\ell}^{\prime}, & & \ell=1 \ldots L 
\end{aligned}
\end{equation}
where $L$ denotes the numbers of the transformer encoders, and $\mathbf{z}_{\ell}$ denotes the output of $l$-th encoder.

ViT has shown superior performance in various computer vision tasks such as image classification, object detection, and semantic segmentation. 
Due to the flexibility of token operation in ViT, which facilitates cross-modal information fusion and interaction, we use ViT as the feature encoder for extracting features from RGB and Sobel modalities.

\subsection{Classification Attentive Regions Proposal}
\label{carp}
The seed regions of CAR maps are located in a weakly supervised learning manner: given class labels, the forgery detection model classifies the faces and generates CAR maps jointly. 
To comprehensively locate forgery cues in the forged faces, we carefully designed the CARP module.

Let the training set be composed of pairs of labels and images, denoted as $\{(\mathbf{x_{i}}, y_{i})\}_{i=1}$ , where $\mathbf{x_i}\in {R}^{H \times W \times 3}$ is an image, and $y_{i}\in\{0,1\}$ is the corresponding label, \ie, 0 as real and 1 as fake. 
As shown in Fig. \ref{fig2}, each image is divided into $h \times w$ patches, and each patch is flattened and linearly projected to a token $\mathbf{z}^0_n\in{R^D, n=1,2,\dots,N} $, where $N=h \times w$ is the number of patches and $D$ is the embedding dimension.  
The tokens are then added to position encodings and fed to $L$ cascaded transformer encoder blocks for feature extraction. 

After the feature extraction, we aim to locate forgery cues.
Previous ViT-based weakly supervised learning works usually take advantage of the self-attention maps in the last transformer block for location proposal \cite{gao2021ts,ru2022learning}, which is intuitive. 
However, the softmax operation in the self-attention mechanism forces tokens to compete with each other for classification.
This works well for multi-class datasets containing similar categories, such as \textit{Sooty Albatross} versus \textit{Laysan Albatross} in CUB-200-2011 \cite{wah2011caltech}, and \textit{terrapin} versus \textit{mud turtle} in ILSVRC \cite{deng2009imagenet}, where discriminating between similar categories requires focusing on more than one area.
Thus, the \textit{winner-take-all} effect is relieved in these works.
However, in face forgery detection, only the most obvious forged region is located as one strong cue is discriminative enough to classify a forged face.
Such concentration impedes the face forgery detection model from learning richer cues.
Additionally, training a ViT through optimizing the class token makes the image tokens class-generic \cite{gao2021ts}, which means that the attentive region contributing to the opposite class is not suppressed in the binary classification procedure.

\begin{figure}[tp]
\centering
{\includegraphics[width=\linewidth]{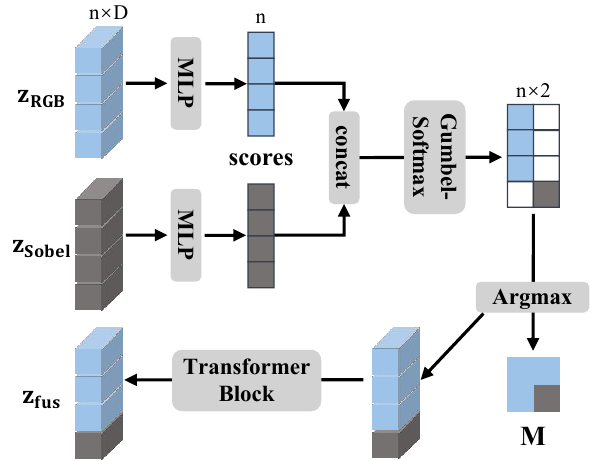}
\caption{Diagrams of the Complementary Learning block. The argmax operation is implemented by matrix production between hard Gumbel-Softmax logits and concatenated tokens. Best view in color.}
\label{fig3}
}
\end{figure}

Instead of directly using the attention map, we sidestep the issues mentioned above by discarding the attention weights of class token and adopting image tokens to propose CAR maps.
Toward this goal, we generate CAM \cite{zhou2016learning} maps in a fully convolutional block specially designed for forgery cue localization.
The output tokens $\mathbf{z} \in{R^{n\times D}}$ are aggregated to feature maps $\mathbf{f}\in{R^{h\times w\times D}}$ and separated along the channel dimension to $\mathbf{f'}\in R^{(h \times w \times d)\times2}$ with a $1\times1$ convolutional layer, where $d$ is the channels for both classes and 2 is the number of classes. 
By separating the feature maps and assigning $d$ maps to both classes, we aim to capture $d$ discriminative area, \ie, one discriminative area in one map, for each class.
We set $d$ as 32 to attend to no more than 32 patches.
$\mathbf{f'}$ is then pooled and softmaxed to $\mathbf{\hat{y}}_{\rm loc}\in R^{2}$ with spatial-wise max-pooling and channel-wise average-pooling.
As elements in $\mathbf{\hat{y}}_{\rm loc}$ are the average value of the maximum in feature maps of both classes, we adopt cross-entropy loss to enlarge the elements corresponding to the ground truth class.
In this way, given the ground truth class $c$, the response values of the most discriminative patches in the corresponding feature maps $\mathbf{f'}_{\rm c}$ are enlarged.
Thus, using CAM to generate CAR maps can locate discriminative areas for richer forgery cues and sidestep the non-maximum suppression problem in the self-attention mechanism in ViT.
Besides, the activated area of the opposite class is suppressed in CAM, constraining the located areas to forgery cues.
The CAR map $\boldsymbol{a}$ is derived as:
\begin{equation}\label{mapgen}
\boldsymbol{a}=\mathrm{sigmoid}\left(\sum_{\rm c~\in~\{0, 1\}} \mathbf{\hat{y}}_{\rm loc_c} \times \boldsymbol{f}_{avg}\left(\mathbf{f}_{\rm c}^\prime\right)\right), 
\end{equation}
where $\rm c$ is class, $\boldsymbol{f}_{avg}$ is spatial-wise average, and $\times$ is scalar multiplication of tensors.
The cross-entropy loss is named localization loss $\mathcal{L}_{\rm loc}$ as it is used to locate forgery cues.


\subsection{Complementary Learning}

To enhance the CAR maps with edge-related forgery cues, we propose to use Sobel-filtered images as a complementary modality.
The Sobel-filtered images are inputted to a network branch identical to the RGB branch.
In these two branches, the extracted tokens $\mathbf{z}$, the aggregated feature map $\mathbf{f}$, and the CAR map $\boldsymbol{a}$ are annotated with RGB and Sobel as subscripts as shown in Fig. \ref{fig2}.
To fully harness the complementary relationship of forgery cues between RGB and Sobel images, we introduce a Complementary Learning (CL) block for cross-modal fusion.
The images from both modalities are first fed to two separate ViTs for inner-modal feature extraction and then fed to the CL block for cross-modal fusion.

The architecture of the CL block is illustrated in Fig. \ref{fig3}.
The CL block dynamically substitutes RGB tokens with Sobel tokens for further classification learning.
Since the RGB and Sobel modalities of the image are homogeneous, an RGB token can be conveniently substituted with a Sobel token of the same spatial position.
Ideally, less discriminative RGB tokens are replaced for better classification performance.
To realize the dynamic substitution across modalities, the scores that determine the importance of each token are required. 
Based on such scores, the CL block selects and discards tokens from two different modalities of the same position.
Therefore, the CL block predicts the importance scores of tokens using two MLPs and substitute tokens accordingly.
This design is particularly suitable for homogeneous modalities, \ie, RGB and Sobel, to interact with each other. 
To enable alignment of the two modalities during training and to make the training process more stable, we use fixed positional encoding in the ViT for both modalities.

However, there is an obstacle that needs to be overcome to perform token substitution.
The obstacle comes due to the network tending to mine features for classification in a single modality as the features of the RGB and Sobel modalities are not aligned in the early training stage. 
This misalignment makes Cross-modal Fusion challenging, which impedes CL from mining forgery cues across modalities.
Subsequently, the CL block tends to predominantly use one single modality, which leads to a trivial solution.
Thus, we introduce competition between RGB and Sobel tokens of the same spatial position to overcome the collapse.
To filter out relatively uninformative tokens, the scores of the tokens are concatenated and then passed through a softmax function.
After, an uninformative RGB token is replaced with the Sobel token of the same spatial position by an argmax operator.

\begin{algorithm}[H]
\caption{PyTorch-like pseudo-code for CL.}
\label{CMF}

\definecolor{codeblue}{rgb}{0.25,0.5,0.5}
\lstset{
  backgroundcolor=\color{white},
  basicstyle=\fontsize{7.2pt}{7.2pt}\ttfamily\selectfont,
  columns=fullflexible,
  breaklines=true,
  captionpos=b,
  commentstyle=\fontsize{7.2pt}{7.2pt}\color{codeblue},
  keywordstyle=\fontsize{7.2pt}{7.2pt},
}
\begin{lstlisting}[language=python]
# MLP_0, MLP_1: score predictors for RGB and Sobel tokens
# z_RGB, z_Sobel: RGB and Sobel tokens (NxD)
# a_RGB, a_Sobel: RGB and Sobel CAR maps (hxwxD)
# N: N=hxw, the number of patches
# tau: temperature for softmax, set to 1

# scores_RGB, scores_Sobel: Nx1
scores_RGB, scores_Sobel = MLP_0(z_RGB), MLP_1(z_Sobel)
# logits: Nx2
logits = cat([scores_RGB, scores_Sobel]).softmax(-1)
gumbel_dist = Gumbel(0, 1)
gumbels = gumbel_dist.sample(logits.shape)
noisy_logits = (logits + gumbels) / tau  
# y_soft: Nx2
y_soft = noisy_logits.softmax(-1)
index = y_soft.max(-1, keepdim=True)[1]
# y_hard: Nx2
y_hard = zeros_like(logits).scatter_(dim, index, 1.0)
# onehot: Nx2
# reparameterization
onehot = y_hard - y_soft.detach() + y_soft
# binary complementary mask
mask = onehot.view(h, w, 2)
# fused tokens
z_fus = mask[:,:,0] * z_RGB + mask[:,:,1] * z_Sobel
# fusion CAR map
a_fus = mask[:,:,0] * a_RGB + mask[:,:,1] * a_Sobel
\end{lstlisting}
\end{algorithm}

As gradients are only backpropagated through the tokens selected by the argmax operation, we adopt the Gumbel-Softmax trick \cite{gumbel} for fully end-to-end training. 
This technique facilitates backpropagation of the gradient through the sampling process, enabling end-to-end learning of the entire network.
Thus, using Gumbel-Softmax is crucial for the Complementary Learning module to learn cross-modal features.
Specifically, it involves adding Gumbel noise to the logits of the categorical distribution, applying softmax to these noisy logits, and subsequently employing an argmax operator for the selection process.
The addition of Gumbel noise introduces an element of randomness to the sampling procedure, which prevents the process from always selecting the larger value.
We further use the reparameterization trick \cite{reparam} to make the argmax operator differentiable.

After substitution, the fused tokens are fed into a transformer block and then aggregated for further classification learning, which average-pool and linear-project the output fused tokens to $\mathbf{\hat{y}}_{\rm fus}$.
We use a cross-entropy loss $\mathcal{L}_{\rm fus}$ to optimize the fused tokens through $\mathbf{\hat{y}}_{\rm fus}$.
During training, the RGB tokens with weaker forgery cues are substituted with Sobel tokens at the same spatial position in the CL block for better classification performance.
Thus, the binary complementary mask $\mathbf{M}$ generated by the argmax operation enables cross-modal forgery cue fusion. 
The Fusion CAR map is derived as:
\begin{equation}
 \boldsymbol{a}_{\rm fus} = \mathbf{M} \odot \boldsymbol{a}_{\rm RGB} + (\mathbf{1}-\mathbf{M}) \odot \boldsymbol{a}_{\rm Sobel} ,
 \label{fusion_eqn}
\end{equation}
where $\odot$ is Hadamard product.
Algorithm \ref{CMF} provides the pseudo-code of Cross-Modal Fusion in a PyTorch-like style.

From the perspective of the RGB modality, the forgery cues in CAR map $\boldsymbol{a}_{\rm RGB}$ are expanded with cues $\boldsymbol{a}_{\rm Sobel}$ in the substitution and constrained to exploitable forgery cues in classification.
In this way, the Fusion CAR map $\boldsymbol{a}_{\rm fus}$ captures forgery cues in both modalities.
We adopt the fused CAR maps $\boldsymbol{a}_{\rm fus}$ as manipulation maps for forged faces.
Our maps can serve as auxiliary supervision for multi-task face forgery detection methods which require pixel-level annotations for forgery cues.

\subsection{Loss Function}
The FoCus is optimized with two losses based on cross-entropy loss.
We use localization loss $\mathcal{L}_{\rm loc}$ to locate the regions that are most relevant to classification in two modalities. 
\begin{equation}
\mathcal{L}_{\rm loc}=-\frac{1}{N} \sum_{i=1}^{N}\sum_{m} \sum_{c=0}^{1} \mathbbm{1}\left(y_{i}=c\right) \log \mathbf{\hat{y}}^{m}_{\rm loc_{c}}, 
\end{equation}
where $m$ denotes modality, and $\mathbbm{1}$ is indicator function. 
To fuse tokens from both modalities for classification, $\mathcal{L}_{\rm fus}$ is formulated as:
\begin{equation}\label{total_object}
 \mathcal{L}_{\rm fus}=-\frac{1}{N} \sum_{i=1}^{N} \sum_{c=0}^{1} \mathbbm{1}\left(y_{i}=c\right) \log \mathbf{\hat{y}}_{\rm fus_{c}}, 
\end{equation}
where $\mathbf{\hat{y}}_{\rm fus}$ is the output of the classification head attached to the fused tokens.
The overall object function for the proposed FoCus is derived as:
\begin{equation}
\rm \mathcal{L}=\mathcal{L}_{\rm loc} + \alpha \mathcal{L}_{\rm fus} , 
\end{equation}
and we set $\alpha$ as 0.1 to make $\mathcal{L}_{\rm loc}$ and $\mathcal{L}_{\rm fus}$ converge to the same order of magnitude.

\section{Experiments}
\begin{figure}[tp]
\centering
{\includegraphics[width=\linewidth]{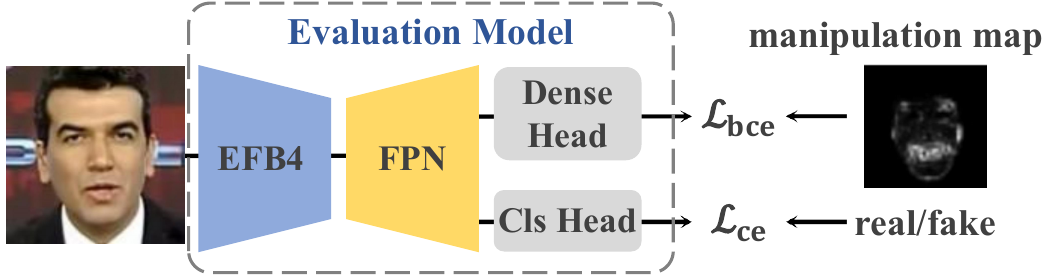}
\caption{The multi-task model for evaluating manipulation maps. Different manipulation maps are used as supervision for the dense head.
We use the classification performance of the evaluation model to assess the exploitability of different manipulation maps.
}
\label{fig4}
}
\end{figure}

In this section, we evaluate the proposed FoCus against other manipulation map generation methods.
As none of the current existing face forgery datasets offer ground truth manipulation maps for evaluation, we  evaluate different manipulation maps through training multi-task face forgery detectors.
We first present a simple but representative multi-task evaluation model for evaluating manipulation maps.
As illustrated in Fig. \ref{fig4}, the evaluation model is built with an EfficientNet-b4 (EFB4) \cite{tan2019efficientnet} backbone and an FPN \cite{lin2017feature} branch to conduct classification and manipulation map prediction, respectively. 
Under different supervision for the predicted manipulation map, the difference in model performance reflects the effectiveness of the manipulation maps generated by different methods. 

\subsection{Experimental Setup}
\noindent\textbf{Datasets.} We conduct experiments on five widely used face forgery datasets, \ie, FaceForensics++ (FF++) \cite{rossler2019faceforensics++}, CelebDeepfake v2 (CDF) \cite{li2020celeb}, DFDC full (DFDC) \cite{dolhansky2020deepfake}, DFDC preview (DFDCp) \cite{dolhansky2020deepfake}, and WildDeepfake (WDF) \cite{zi2020wilddeepfake}. 
\textbf{FF++} contains 4,000 forged videos and 1,000 real videos. There are four forgery techniques used in FF++, including two for identity swap (DF and FS) and two for face reenactment (NT and F2F). All videos are provided in three compression levels: raw, high-quality (HQ), and low-quality (LQ). 
\textbf{CDF} contains 590 real videos and 5,639 forged videos corresponding to 59 celebrities. The forged videos are generated using an improved face swap algorithm \cite{li2020celeb}.  \textbf{DFDC} is a large-scale dataset that contains more than 120,000 video clips of 486 subjects filmed in extreme conditions such as large poses and low lighting. 
The forged videos were generated using eight forgery techniques.
\textbf{DFDCp} is the preview version of DFDC and contains about 5,000 videos. 
\textbf{WDF} contains 3,805 real and 3,509 forged video clips collected from the internet. The video clips in WDF are diverse in identities and manipulation methods.
We use FF++ for training and testing, while the other four datasets are for generalization evaluation. 
For DFDC, we use videos that contain one subject for testing to avoid labeling noise considering the presence of multiple subjects in a video. 
For other datasets, we follow their official dataset splits for testing. 

\noindent \textbf{Compared methods.} We compare our method with two commonly used comparison-based manipulation map generation methods, \ie, SSIM and pixel-wise difference. 
Previous works often employ a threshold of 0.1 to filter out noises when using pixel-wise difference \cite{lttd}. 
Therefore, we also include a comparison with this strategy which we refer to as Pixel-wise Different@0.1. 
We also compare the performance of the evaluation model supervised by the spatial attention maps in three face forgery detectors, \ie, MaDD \cite{zhao2021multi}, GFFD \cite{luo2021generalizing}, and UIA-ViT \cite{uiavit}. 
These methods locate forged regions within their attention maps. 
We note that these methods achieved \textit{sota} performance as they are capable of locating forgery cues.
Now, we compare the manipulation maps generated by FoCus with those generated by these methods in terms of locating forgery cues through the following evaluation method.

\noindent \textbf{Evaluation protocol.} The proposed FoCus is trained on the FF++ to generate fusion CAR maps. 
Then, we modify the CAR maps to manipulation maps by setting the CAR maps of real faces to all-zero maps. 
As shown in Fig. \ref{fig4}, manipulation maps generated by different approaches are used as auxiliary supervision to train a simple multi-task evaluation model.
We use cross-entropy loss $\mathcal{L}_{ce}$ and binary cross-entropy loss $\mathcal{L}_{bce}$ for classification and map prediction tasks to optimize the evaluation model. 
The scale factor between these two losses is 0.1.
We report its Accuracy (Acc) score and Area Under the Receiver Operating Characteristic Curve (AUC) score in the in-dataset evaluation on FF++, and AUC scores in the cross-dataset evaluation on the other four datasets.
All scores reported are frame-level results.

\noindent \textbf{Implementation.} We sample 10 frames per video and use RetinaFace \cite{deng2020retinaface} to crop faces. A pre-trained ViT small is adopted \cite{dosovitskiy2020image} as the backbone for FoCus. 
The faces are resized to $224\times224$. The size of the output $\boldsymbol{a}_{\rm fus}$ is $14\times14$. 
The evaluation model outputs the predicted map in size $40\times40$.
We employ Bilinear interpolation to resize it to $40\times40$ and set $\boldsymbol{a}_{\rm fus}$ for real faces to all zeros as supervision.
FoCus is trained using an Adam optimizer with a learning rate of 1e-4 and a batch size of 128 for 30,000 iterations.
We use cosine decay as the learning rate schedule.

\subsection{In-dataset and Cross-dataset Evaluation}
\begin{table}[t]
\caption{In-dataset Acc(\%) and AUC(\%) of evaluation model supervised by different maps. Bold indicates the best results.}
\label{tab1}
\centering
\resizebox*{0.86\linewidth}{!}{
\begin{tabular}{ccccc}
\hline
\multirow{2}{*}{\begin{tabular}[c]{@{}c@{}}Map Generation \\ Method\end{tabular}} & \multicolumn{2}{c}{FF++(HQ)} & \multicolumn{2}{c}{FF++(LQ)} \\ 
& Acc   & AUC    & Acc   & AUC    \\ \hline
SSIM& 94.34 & 98.39 & 81.83 & 87.66 \\
Pixel-wise& 94.64 & 98.35 & 80.91 & 87.37 \\
Pixel-wise@0.1&  94.69&  98.75&  83.31& 88.16 \\
GFFD \cite{luo2021generalizing} & 95.51 & 98.81 & 84.34 & 88.98 \\
MaDD \cite{zhao2021multi}& 95.54 & 98.69 & 82.86 & 87.76 \\
UIA-ViT \cite{uiavit}& 96.06&98.97  &86.71  &89.62  \\
FoCus(Ours)          & \textbf{96.43}      &  \textbf{99.15}      & \textbf{87.31}      & \textbf{91.01}       \\ \hline
\end{tabular}
}
\end{table}

Through quantitative experiments, we verify using manipulation maps generated by FoCus as auxiliary supervision can efficiently improve the performance of multi-task face forgery detection models.
Specifically, we adopt in-dataset and cross-dataset evaluation protocols for evaluating multi-task models supervised by different manipulation maps.

\noindent \textbf{In-dataset evaluation.} We first evaluate the evaluation model in Fig. \ref{fig4} on the FF++ dataset under LQ and HQ settings.
As shown in Table~\ref{tab1}, using the maps generated by FoCus as supervision outperforms all compared methods. 
The AUC scores of utilizing our maps exceed the best results among other methods by 1.39\% and 0.18\% under LQ and HQ.
Moreover, our FoCus consistently surpasses top results among other methods under LQ and HQ in improving the Acc scores of the evaluation model. 
The in-dataset evaluation illustrates that using the maps generated by FoCus as supervision is superior in guiding the model to learn more discriminative forgery cues. 
It also verifies the proposed FoCus can locate more exploitable forgery cues.

\noindent \textbf{Cross-dataset evaluation.} We conduct cross-dataset experiments to investigate the impact of different manipulation maps used as supervision on the generalizability of the evaluation model.
As shown in Table~\ref{tab2}, the evaluation model supervised by our maps maintains better generalizability than all competitors in cross-dataset evaluation.
Compared with the top results of other methods, using our maps as auxiliary supervision improves the AUC by 4.01\% and 0.99\% when generalizing to the CDF dataset after training on FF++(HQ) and FF++(LQ).
The performance gain is consistently significant when testing on datasets generated by various face forgery methods such as DFDC. 
On average, using our maps as supervision improves the AUC scores by 1.45\% on DFDC, 2.62\% on DFDCp, and 2.10\% on WDF after training on FF++(HQ) and FF++(LQ). 

\begin{table}[t]
\caption{Cross-dataset AUC of evaluation model supervised by different maps. }
\resizebox*{\linewidth}{!}{
\begin{tabular}{cccccc}
\hline
\multirow{2}{*}{\begin{tabular}[c]{@{}c@{}}Training\\ Dataset\end{tabular}} 
& \multirow{2}{*}{\begin{tabular}[c]{@{}c@{}}Map Generation\\Method\end{tabular}} & \multicolumn{4}{c}{Testing Dataset} \\ \cline{3-6} 

         &       & CDF  & WDF & DFDC  & DFDCp    \\ \hline
\multirow{7}{*}{\begin{tabular}[c]{@{}c@{}}FF++\\(HQ)\end{tabular}}       
         & SSIM         & 72.12     & 72.38 & 65.35&72.92 \\
         & Pixel-wise   & 69.71     & 71.59 & 66.06&69.37 \\
         & Pixel-wise@0.1 & 69.47    & 69.29  & 67.64&70.31\\
         & GFFD         & 68.37     & 69.17 & 64.15&67.94 \\
         & MaDD         & 68.07     & 68.16 & 61.95&70.82 \\
         & UIA-ViT      & 66.73     & 71.04 & 67.16&73.56\\
& FoCus(Ours) &\textbf{76.13} &    \textbf{73.31}   &    \textbf{68.42}  & \textbf{76.62}      \\ \hline
\multirow{7}{*}{\begin{tabular}[c]{@{}c@{}}FF++\\(LQ)\end{tabular}}       
& SSIM         & 68.79   & 68.95  & 64.71 & 71.59\\ 
         & Pixel-wise & 70.22     & 66.94  & 63.76&70.21\\
         & Pixel-wise@0.1 & 70.55    & 68.35  & 64.80&72.98\\
         & GFFD       & 71.03  & 69.31   & 62.07&  70.81\\ 
          & MaDD       & 70.27     & 68.04  & 61.94 & 69.17\\
          & UIA-ViT       & 64.60    & 70.26   & 65.29 &75.63\\
& FoCus(Ours)    & \textbf{72.02}  & \textbf{72.18}  & \textbf{66.92}   &  \textbf{77.80}   \\ \hline
\end{tabular}
}

\label{tab2}
\end{table}

\noindent \textbf{Evaluation on more models.}
To validate the effectiveness of the proposed FoCus in supervising existing face forgery detection methods, we reproduce PCL\cite{zhao2021learning}, LTTD \cite{lttd}, and CADDM\cite{dong2023implicit}.
Each of these face forgery detectors originally includes a segmentation training head. 
For PCL and CADDM, we use their publicly released codes. 
For LTTD, we reproduce it following the details in the paper and supplementary.
PCL generates manipulation maps with SSIM and attaches a patch-level metric learning branch to a ResNet34 backbone.
LTTD is a ViViT-like\cite{vivit} video-level model that models the temporal features of video patch sequences. 
It introduced a branch for predicting the integrity of patch sequences, with annotations derived from pixel-wise differences between real and fake videos.
CADDM extracts local artifact areas on images to pay less attention to the global identity to avoid identity leakage.
We replace their auxiliary supervision with manipulation maps of FoCus and evaluate the performance for comparison.
For PCL, we omit their data augmentation scheme.

The results are listed in Table~\ref{tab3}. 
It is obvious that CADDM gains better performance when supervised by our maps.
For instance, the in-dataset AUC of CADDM improved by 0.95\% using our maps as auxiliary supervisions for segmentation training in average.
Moreover, the improved performance of PCL and LTTD verifies the effectiveness of our maps in supervising patch-level metric learning and video-level models.
To sum up, FoCus consistently surpasses comparison-based methods and attention maps in previous detectors in supervising four multi-task models and improves their performance by a large margin. 
This verifies that FoCus captures more general and accurate forgery cues than existing methods.

\begin{table*}[t]
\caption{In-dataset and cross-dataset performance of models in \cite{zhao2021learning,lttd,dong2023implicit} supervised by different maps. We replace their map generation methods with our FoCus for comparison. }
\label{tab3}
\centering
\resizebox*{0.75\linewidth}{!}{%
\begin{tabular}{ccccccccc}
\hline
\multirow{3}{*}{Model} &
\multirow{3}{*}{\begin{tabular}[c]{@{}c@{}}Training \\Dataset\end{tabular}} &
\multirow{3}{*}{\begin{tabular}[c]{@{}c@{}}Map \\Generation\\Method\end{tabular}} &
\multicolumn{6}{c}{Testing Dataset} \\ \cline{4-9} 
&      &     & \multicolumn{2}{c}{FF++} & CDF & WDF & DFDC & DFDCp  \\
&      &     & Acc & AUC  & AUC      & AUC    & AUC & AUC  \\ \hline
\multirow{4}{*}{\begin{tabular}[c]{@{}c@{}}PCL\\\cite{zhao2021learning}\end{tabular}} &
\multirow{2}{*}{\begin{tabular}[c]{@{}c@{}}FF++(HQ)\end{tabular}} &
SSIM & 94.21 & 96.23 & \textbf{78.46} & 67.46 & 66.75 &70.14\\
&      & FoCus(Ours) & \textbf{95.50}      & \textbf{96.89}      & 76.77   & \textbf{72.42} & \textbf{68.11} &\textbf{74.22} \\ \cline{2-9} 
& \multirow{2}{*}{\begin{tabular}[c]{@{}c@{}}FF++(LQ)\end{tabular}} & SSIM   & 83.70      & 87.84      & \textbf{73.01}   & 67.90 & 62.83  &69.15\\
&      & FoCus(Ours) & \textbf{84.12}   & \textbf{88.02}      & 71.32   & \textbf{71.75} & \textbf{66.79} &\textbf{73.25}\\ \hline
\multirow{4}{*}{\begin{tabular}[c]{@{}c@{}}LTTD 
\\ \cite{lttd}\end{tabular}} &
\multirow{2}{*}{\begin{tabular}[c]{@{}c@{}}FF++(HQ)\end{tabular}} &
Pixel-wise & \textbf{97.47} & 98.61 & 78.67 & 73.19 & 70.90 &76.33\\
&     & FoCus(Ours) & 97.08      & \textbf{99.12}      & \textbf{79.38}   & \textbf{76.60} & \textbf{74.95} & \textbf{79.83}\\ \cline{2-9} 
& \multirow{2}{*}{\begin{tabular}[c]{@{}c@{}}FF++(LQ)\end{tabular}} & Pixel-wise   & 89.27      & 92.65      & 80.57   & 73.36 & 69.27 &74.06 \\
&      & FoCus(Ours) & \textbf{89.50}   & \textbf{93.19}      & \textbf{81.80}   & \textbf{76.83} & \textbf{71.84} &\textbf{76.62}\\\hline
\multirow{4}{*}{\begin{tabular}[c]{@{}c@{}}CADDM 
\\ \cite{dong2023implicit}\end{tabular}} &
\multirow{2}{*}{\begin{tabular}[c]{@{}c@{}}FF++(HQ)\end{tabular}} &
SSIM & 94.23 & 96.12 & \textbf{83.20} & 78.99 & \textbf{73.27} & 78.50\\
&     & FoCus(Ours) &  \textbf{94.62} & \textbf{97.17} & 82.84 & \textbf{80.75} &  72.81 & \textbf{80.39}\\ \cline{2-9} 
& \multirow{2}{*}{\begin{tabular}[c]{@{}c@{}}FF++(LQ)\end{tabular}} & SSIM   & 88.16     & 91.23     & 81.15   &78.69  & 71.81 & 75.44 \\
&      & FoCus(Ours) & \textbf{90.30}   & \textbf{92.08}      & \textbf{82.92}   & \textbf{79.80} & \textbf{72.96} &\textbf{79.86}\\\hline
\end{tabular}
}
\end{table*}

\begin{figure}[tp]
\centering
{\includegraphics[width=\linewidth]{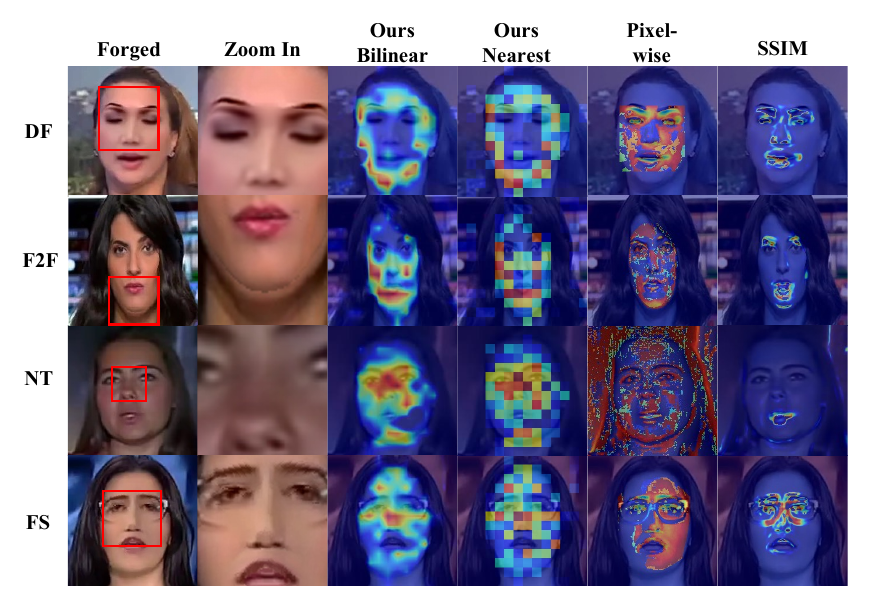}
\caption{The generated maps for fake faces in FF++(HQ). We provided maps interpolated through Bilinear and Nearest interpolation for better visualization.}
\label{fig_vis}
}
\end{figure}

\begin{figure}[]
\centering
{\includegraphics[width=0.88\linewidth]{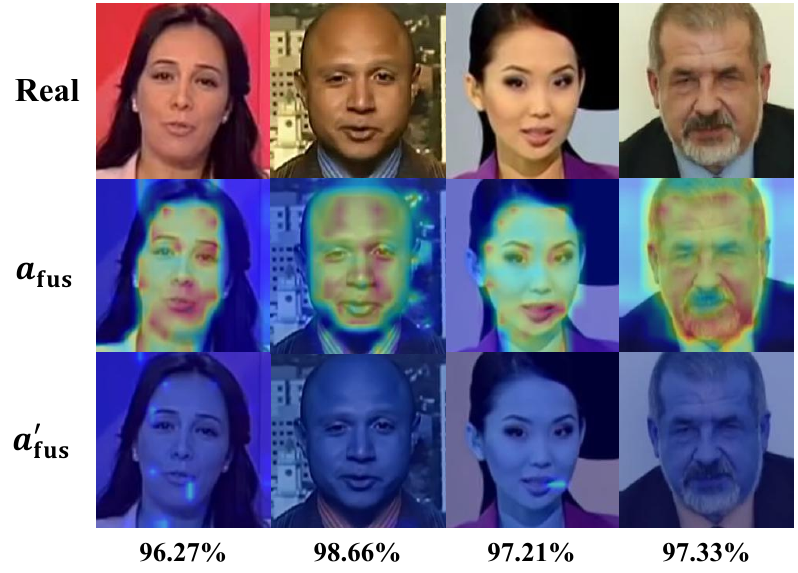}
\caption{Visualization of generated maps for real faces in FF++. The second row is $\boldsymbol{a}_{\rm fus}$ in Equation \ref{mapgen}, which indicates that FoCus searches throughout the entire face to determine a face is real. The third row is $\boldsymbol{a}^{\prime}_{\rm fus}$ in Equation \ref{modified_mapgen}, which indicates that FoCus rarely finds forgery cues in real faces. The bottom are the predicted probabilities of being real by $\mathbf{\hat{y}}_{\rm fus}$.
\label{realfaces}
}
}
\end{figure}

\subsection{Visualization}
\noindent \textbf{Generated maps for fake faces.}
We first visualize the maps generated by FoCus along with pixel-wise difference and SSIM in Fig. \ref{fig_vis} to demonstrate their ability to locate forgery cues. 
Compared with comparison-based methods that mark all changed pixels, FoCus is superior in locating apparent anomalous forgery cues.
For example, FoCus accurately locates seam-like artifacts around the blending boundary in DF and FS. 
Artifacts such as multiple eyebrows and incomplete glasses are also accurately highlighted by FoCus.
In particular, FoCus locates exploitable cues around the swapped faces generated through DF, instead of the entire inner face.
Such annotations are more efficient in guiding detectors to learn forgery cues and are highly interpretable.
In contrast, pixel-wise difference locates the inner part of the face, which is locally normal.
Forcing detectors to locate such areas impedes model learning stitching artifacts.
Thus, multi-task models supervised by our maps have better generalizability.
For NT, our FoCus locates obscure areas that SSIM fails to find. 
Besides, as NT contains a decoder for image synthesizing, changed pixels are added to backgrounds. 
These changed pixels are annotated as forged by pixel-wise difference, resulting in poor interpretability and exploitability.

\noindent \textbf{Generated maps for real faces.}
We visualize the $\boldsymbol{a}_{\rm fus}$ generated by our FoCus for real faces in Fig. \ref{fig_vis}. 
As stated before, the manipulation maps essentially indicate the regions used to classify the image into the category predicted by the network. 
Therefore, for real faces, $\boldsymbol{a}_{\rm fus}$ shows which areas the model has examined to classify the face as real and does not imply that FoCus detects false positives in real faces.
As shown in the second row of Fig \ref{realfaces}, FoCus scans the entire face for anomalies to determine whether a face is real. 
In previous in-dataset and cross-dataset evaluations, the manipulation maps for real faces in training multi-task models are set to all zero as the binary classification labels are known.
Therefore, the located areas in $\boldsymbol{a}_{\rm fus}$ will not interfere with the training of the multi-task face forgery detection model.

\begin{figure*}[]
\centering
{\includegraphics[width=0.95\textwidth]{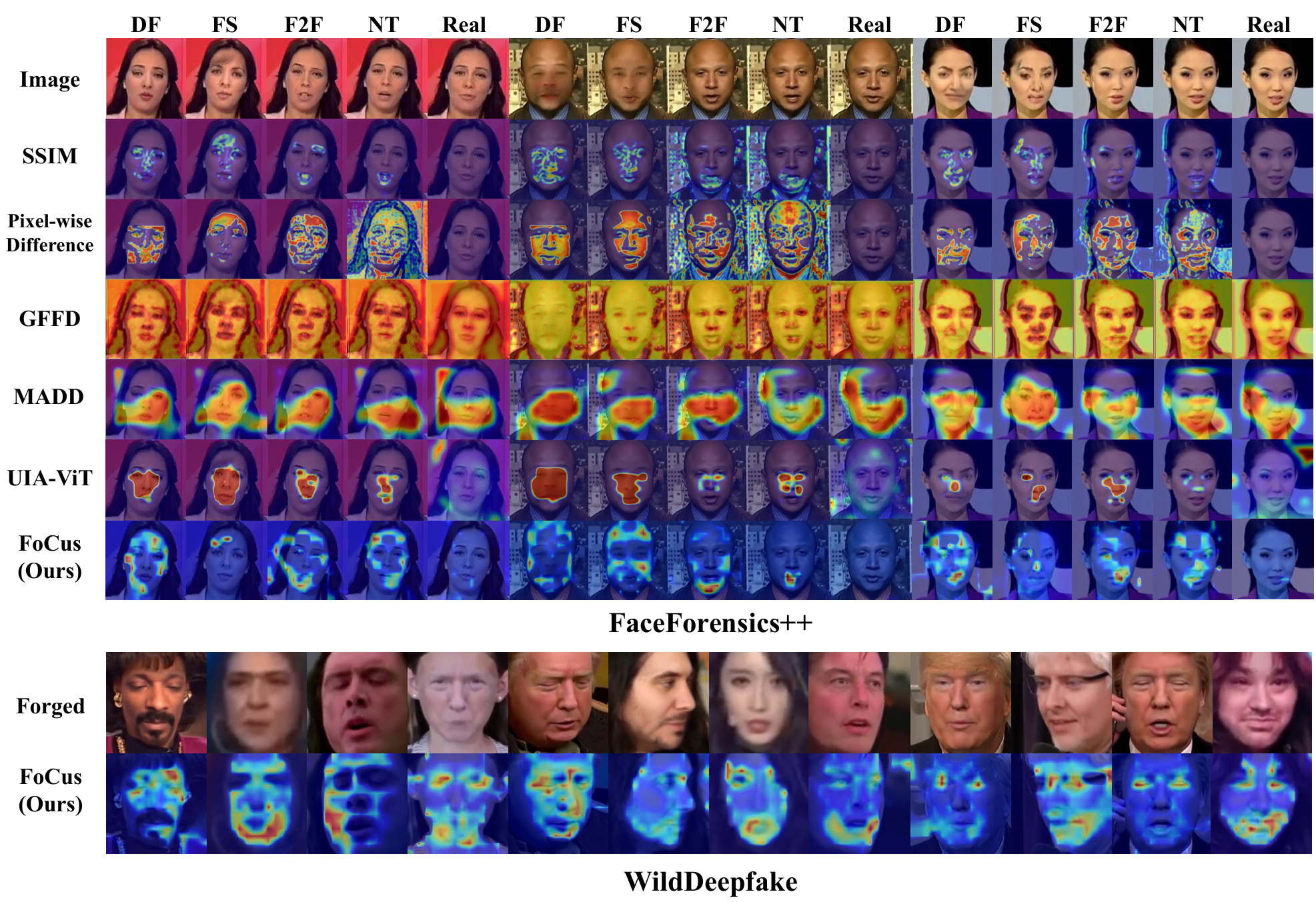}
\caption{Visualization of generated maps for faces in FF++ and WDF. As fake faces in FF++ have corresponding real faces, we also visualize maps for real faces in FF++ for comparison.}
\label{sup1}
}
\end{figure*}

To further investigate whether false detection occurs in real faces, one can simply modify the subscript in Equation \ref{mapgen} in the CARP module to: 
\begin{equation}\label{modified_mapgen}
\boldsymbol{a}^{\prime}=\operatorname{sigmoid}\left(\hat{\mathbf{y}}_{\mathrm{loc}_1} \times \boldsymbol{f}_{avg}\left(\mathbf{f}_1^{\prime}\right)\right).
\end{equation}
The $\boldsymbol{a}^{\prime}$ only focuses on the regions in real faces that contribute to the classification to label 1, \ie, the fake category.
We note the corresponding fusion CAR map as $\boldsymbol{a}^{\prime}_{\rm fus}$.
The visualization results for $\boldsymbol{a}^{\prime}_{\rm fus}$ are shown in the third row of Fig. \ref{realfaces}.  
It is obvious that FoCus rarely finds forgery cues in real faces, \ie, false detection.
When using our FoCus to locate forgery cues in an unlabeled face, one can directly use  $\mathbf{\hat{y}}_{\rm fus}$ to assess whether the face is real and further use Equation \ref{modified_mapgen} to locate the areas contributing to the classification to the fake category.

The difference between Equation \ref{modified_mapgen} and \ref{mapgen} does not affect the usability of FoCus. 
As $\hat{\mathbf{y}}_{\mathrm{loc}_0}$ for fake faces are quite small, the difference between $\boldsymbol{a}_{\rm fus}$ and $\boldsymbol{a}^{\prime}_{\rm fus}$ for fake faces is negligible. 
Therefore, in the previous experiments, we used $\boldsymbol{a}_{\rm fus}$ as manipulation maps.

\noindent \textbf{Compared with other manipulation maps.}
We provide more visualization results of maps generated by different methods for fake and real faces in Fig. \ref{sup1}.
As we aim to locate forgery cues, we adopted $\boldsymbol{a}^{\prime}_{\rm fus}$ as manipulation maps for real faces.
Compared to the maps of the attention mechanism in GFFD, MaDD, and UIA-ViT, the manipulation maps produced by FoCus exhibit better interpretability.
Specifically, the forgery cues located by MaDD and GFFD are widely scattered, and UIA-ViT mechanically marks the inner faces even for NT and F2F where the inner faces are not largely manipulated.
For instance, in the second column of Fig. \ref{sup1}, the most obvious forgery pattern is blurred hair on the forehead, which can be verified by SSIM and pixel-level difference.
FoCus accurately marks this area as forged while GFFD, MaDD, and UIA-ViT fail.
Their maps deviate from the most evident forgery cues.
Besides, our FoCus rarely found forgery traces in real faces, while GFFD and MaDD wrongly located large area in real faces.

Additionally, the visualization for faces in WDF shows that FoCus maintains the ability to discover cues for wild forgeries with lighting and blur differences, while comparison-based methods cannot be adopted for these unpaired faces.
This demonstrates the potential of applying FoCus to source-free face forgeries for future synthesis algorithms.

\begin{table*}[t]
\caption{Robustness to corruptions in \cite{deeper}, measured by average AUC (\%) across five intensity levels. Bold indicates best results. }
\label{robust}
\centering
\resizebox*{0.85\linewidth}{!}
{
\begin{tabular}{lccccccccc}
\hline
Method               & Clean & Saturation & Contrast & Block & Noise & Blur & Pixel & Compress & Avg  \\ \hline
Xception \cite{rossler2019faceforensics++}     & 99.8                         & 99.3       & 98.6     & \textbf{99.7}  & 53.8  & 60.2 & 74.2  & 62.1     & 78.3 \\
Patch-based \cite{patchforensics}  & \textbf{99.9}                         & 84.3       & 74.2     & 99.2  & 50.0    & 54.4 & 56.7  & 53.4     & 67.5 \\
Face X-ray \cite{li2020face}   & 99.8                         & 97.6       & 88.5     & 99.1  & 49.8  & 63.8 & 88.6  & 55.2     & 77.5 \\
CNN-GRU \cite{cnngru}     & 99.8                         & 99.0         & 98.8     & 97.9  & 47.9  & 71.5 & 86.5  & \textbf{74.5}     & 82.3 \\ \hline

Ours (w.o. Sobel)                 & 99.8                         & 99.2       & 99.0     & 98.4    & 60.8 & 79.0 & 90.9  & 66.7     & 84.8 \\
Ours                 & \textbf{99.9}                         & \textbf{99.5}       & \textbf{99.2}     & 98.0    & \textbf{76.5} & \textbf{89.2} & \textbf{97.4}  & 68.6     & \textbf{89.8} \\ \hline
\end{tabular}
}
\end{table*}

\subsection{Robustness}

To ensure the effectiveness of face forgery detectors, it is important that they exhibit robustness against common corruptions and obstructions that face forgery media may encounter on social network sites.
We evaluate the robustness of our approach to perturbations using the protocol proposed in \cite{deeper}.
Further, we use faces with occlusions, extreme expressions, and large poses in the test set of WDF to evaluate the robustness of our approach to these unseen cases.

\noindent\textbf{Common corruptions.}
The image perturbations proposed in \cite{deeper} include variations in contrast and saturation, block-wise occlusions, Gaussian noise and blur, pixelation, and video compression, each with five different intensity levels.

\begin{figure}[t]
\centering
\includegraphics[width=0.9\linewidth]{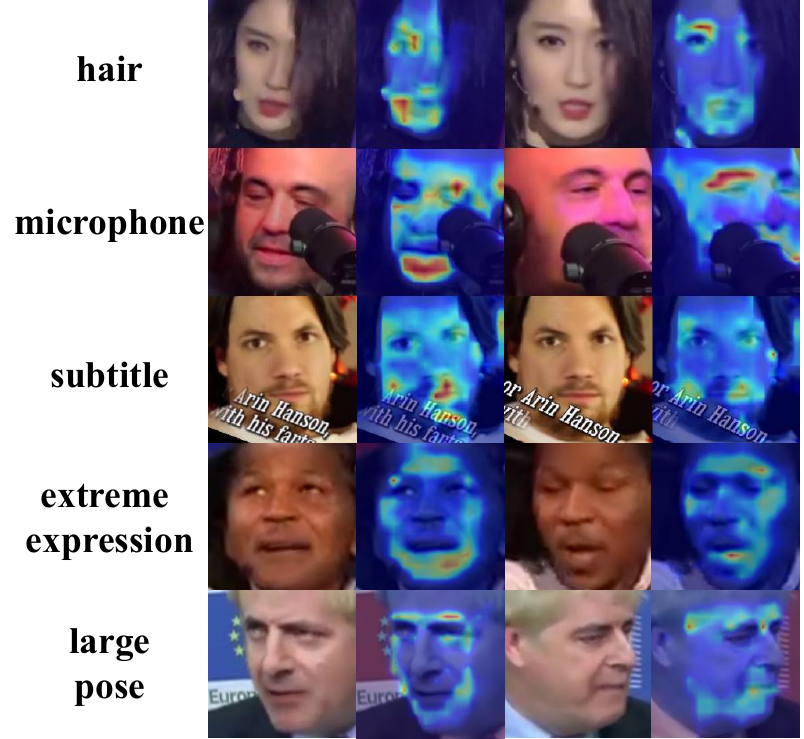}
\caption{Visualization of the manipulation maps of FoCus for faces with occlusions (left) and extreme expressions (right). } 
\label{overlay}
\end{figure}
Table \ref{robust} shows the average AUC across all intensity levels for each corruption type.
The model supervised by our FoCus proves to be more robust against common corruptions compared to other models.
We conjecture that the Sobel inputs induce the model to focus on edge-related forgery patterns, and these traces are more robust against Gaussian blur, noise, and pixelation.
However, since compression tends to degrade the edge details of an image, the robustness of our method against compression is inferior to the video-level detection method CNN-GRU.
CNN-GRU relies on temporal consistency to classify a video clip.
As temporal consistency is robust against compression, its robustness is superior to other frame-level face forgery detection methods in terms of Compression.
However, FoCus still outperforms other frame-level models in the unseen compression.

\noindent\textbf{Occlusions, large poses, and extreme expressions.}
As for the robustness against occlusions, we take advantage of forgeries in the WDF dataset for case studies.
In the test set of WDF, we manually selected partially occluded faces, and there are three videos of them.
They are occluded by subtitles, hairs, and microphones, respectively.
We also investigate the robustness against large poses and extreme expressions.
We use face recognizers in \cite{Face.evo,frwild} to select faces with large poses and extreme expressions.
To demonstrate the effectiveness of our FoCus against these perturbances, we visualize their manipulation maps generated by FoCus in Fig. \ref{overlay}.
As illustrated in Fig. \ref{overlay}, FoCus can effectively locate forged cues without being disturbed by occlusions. 
Specifically, FoCus locates anomalies in faces without marking these occlusions as forgery cues.
Moreover, for faces with extreme expressions and large poses, FoCus consistently locates accurate face blending boundaries.
Visualization of the located forged cues for these samples demonstrates the robustness of FoCus against occlusions, large poses, and extreme expressions.

\subsection{Feature distribution}
\begin{figure}[] 
\centering
{\includegraphics[width=\linewidth]{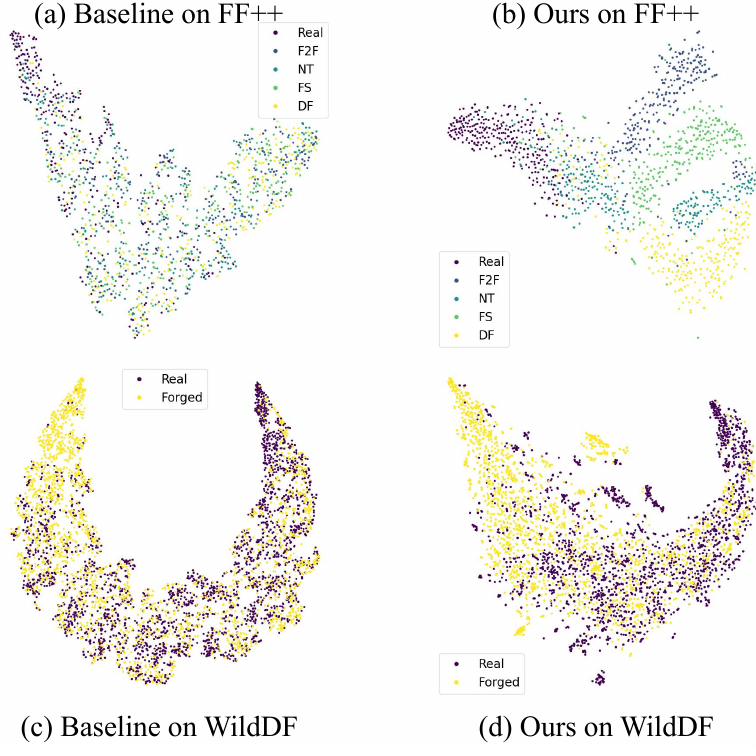}
}\caption{t-SNE of the feature distribution of the model in Fig. \ref{fig4}, trained without supervision (baseline) and trained with our maps. Models are trained on FF++(HQ), and we illustrate the feature distribution on the test set of FF++(HQ) and WDF.}
\label{fig6}
\end{figure}

Our goal is to identify general forgery cues across different types of manipulations and enhance the generalization of detection models to previously unseen forgeries.
By applying our manipulation maps as supervision, the detector is expected to learn more diverse representations for forgery cues.
To investigate how our approach helps to generalize to unseen forgeries, and how discriminative our feature representation is, we visualize the feature distribution of the evaluation models trained with our maps using t-SNE\cite{2008Visualizing}.
The results are presented in Fig. \ref{fig6}, which illustrates the feature distribution of the model on training set FF++ and the previously unseen testing set WDF. 

The model supervised by our maps can automatically recognize different types of forgeries in FF++ and arrange the corresponding samples together clearly in the feature space solely based on the binary labels.
In contrast, the baseline model embeds different types of forgeries to the same cluster in the feature space, which indicates it fails to learn rich forgery cues.
Additionally, we observe that when using our maps as supervision, the intra-class features to previously unseen forgeries in the testing set (WDF) are more dispersed and the inter-class features are more distinguishable.
This suggests that our maps reveal more diverse forgery cues. 
This is due to the fact that our FoCus method provides rich annotations for exploitable forgery cues during the training process, which helps improve the detection models' ability to identify forgeries encountered in various scenarios.
Benefiting from the richer learned forgery cues, the supervised face forgery detection models can better generalize to detect various types of face forgeries.

\begin{table}[t]
\caption{Ablation study for the input modalities. In addition to the Sobel-filtered images, we compare using images filtered by DCT and SRM as auxiliary modalities in FoCus.}\label{tab6}
\resizebox*{\linewidth}{!}
{
\begin{tabular}{ccccccc}
\hline
\multirow{3}{*}{Modalities} & \multicolumn{6}{c}{Testing Dataset}  \\\cline{2-7} & \multicolumn{2}{c}{FF++(LQ)} & CDF & WDF & DFDC &DFDCp   \\  
& Acc   & AUC    & AUC      & AUC   & AUC  & AUC  \\ \hline
RGB    & 85.77   & 88.90 & 71.37   & 70.51   & 62.28 & 73.47 \\\hline
+Sobel (Ours) & \textbf{87.02} & \textbf{91.01}  &\textbf{72.02}     &72.18 	 & \textbf{66.92}  & \textbf{77.80}     \\
+RGB & 86.09 & 89.22  &70.68     & 71.23	 & 63.42  &73.64     \\
+DCT \cite{ahmed1974discrete} & 86.98 & 90.54 &70.66    &\textbf{74.58}  & 64.64 & 72.47     \\
+SRM \cite{fridrich2012rich} & 85.98 &88.02  & 69.67   & 69.86 & 64.15  & 71.40     \\ \hline
\end{tabular}

}
\end{table}

\subsection{Ablation Study} 
\noindent \textbf{Modality.} To demonstrate the effectiveness of introducing Sobel-filtered images for Complementary Learning, we present ablation studies on the complementary modalities, including using Discrete Cosine Transform (DCT) and Steganalysis Rich Model (SRM) to filter images as complementary.
We also evaluate a variant using another RGB branch for comparison. 
It can be observed in Table~\ref{tab6} that the best results are obtained using both RGB and Sobel-filtered images. 
In contrast, using SRM noise as complementary leads to inferior performance.
This indicates that detection models that use RGB modality as inputs struggle to learn subtle forgery cues in the noise view as complementary.
Besides, using DCT as complementary obtains good results on WDF.
From the perspective of data features, this is due to the significant difference in clarity between the inner and outer parts of the fake faces in WDF.
DCT can expose these differences more effectively in high frequency.
Additionally, employing another RGB branch as complementary leads to marginal performance gains, which suggests increasing model capacity aids in exposing richer cues.
Compared to using a Sobel branch as complementary, using another RGB branch is less efficient.

\begin{figure}[tbp]
\centering
{\includegraphics[width=0.85\linewidth]{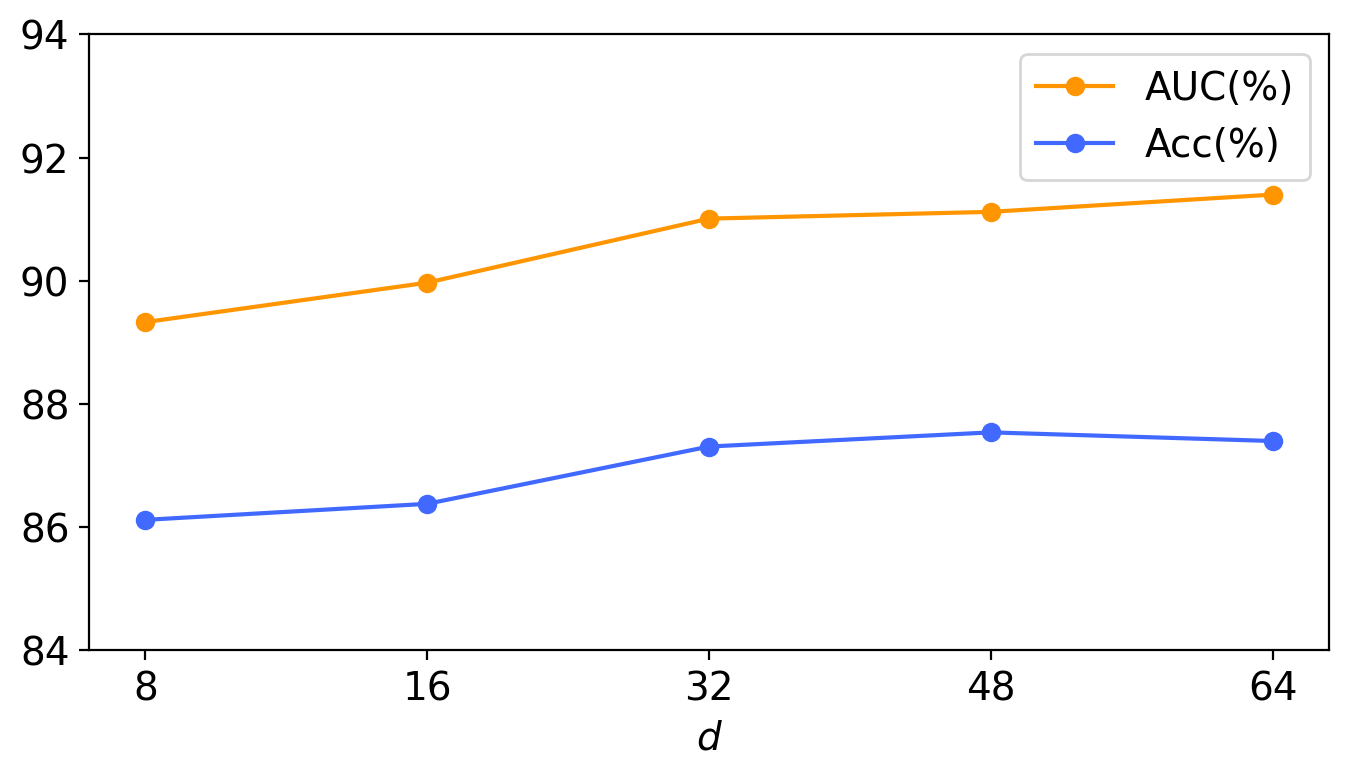}
\caption{The impact on performance by varying $d$. The orange and blue lines show AUC and Acc scores of in-dataset evaluations of evaluation model on FF++(LQ), respectively.}
\label{carp_d}
}
\end{figure}

\begin{table}[t]
\caption{Ablation study for CARP and Complementary Learning (CL) modules. We additionally use ScoreCAM and GradCAM for comparison when both the CARP and CL are not used.}
\label{tab7}
\resizebox*{\linewidth}{!}
{
\begin{tabular}{cccccccc}
\hline
\multirow{3}{*}{Models} & \multicolumn{6}{c}{Testing Dataset}  \\\cline{2-7} & \multicolumn{2}{c}{FF++(LQ)} & CDF & WDF & DFDC & DFDCp  \\  
& Acc   & AUC    & AUC      & AUC   & AUC & AUC    \\ \hline
FoCus     & \textbf{87.02} & \textbf{91.01}  &\textbf{72.02}     &\textbf{72.18} 	 & \textbf{66.92}  & \textbf{77.80}    \\\hline
-CARP & 86.59 & 90.42 & 70.63   & 71.02 & 63.20  & 70.76    \\
-CL & 86.27 & 89.04 & 70.14   & 71.91 & 64.28  & 69.28  \\\hline
GradCAM & 78.13 & 81.35 & 68.44   & 65.19	& 64.29  & 70.04  \\
ScoreCAM & 82.95 & 84.62 & 68.96   & 69.04 & 60.84  & 68.32  \\\hline
\end{tabular}
}
\end{table}

\noindent \textbf{Modules.} 
We perform ablation studies on our CARP module and CL module to demonstrate their effectiveness.
The results are listed in Table~\ref{tab7}.
Four variants are devised for comparison: 1) using attention maps of the self-attention mechanism to serve as CAR maps instead of the proposed CARP module, \ie, -CARP. 
2) using spatial-wise maximum operation to fuse CAR maps in both modalities instead of using the CL module, \ie, -CL. 
3) variant 3 and 4 using two popular CAM variant schemes, \ie, GradCAM \cite{gradcam} and ScoreCAM \cite{scorecam}, to generate CAR maps in a trained ViT small detector, where both CARP and CL modules are not used, \ie, GradCAM and ScoreCAM.

Benefiting from the CAM-like scheme, CARP is not constrained to part of the forgery cues.
As illustrated by the results, adopting CARP for locating forgery cues leads to better performance compared to variant 1(-CARP) which adopts attention maps in the self-attention mechanism.
Besides, compared to variant 2(-CL) which adopts maximum operation, using the proposed CL module for cross-modal fusion leads to flexible forgery cue fusion and better performance.
Furthermore, compared to variant 3 and 4 where two CAM-based schemes are separately used for generating manipulation maps, our FoCus demonstrates better performance.
The experiments for FoCus and its variants verify the effectiveness of the proposed CARP and CL.

\begin{figure}[tbp]
\centering
{\includegraphics[width=0.85\linewidth]{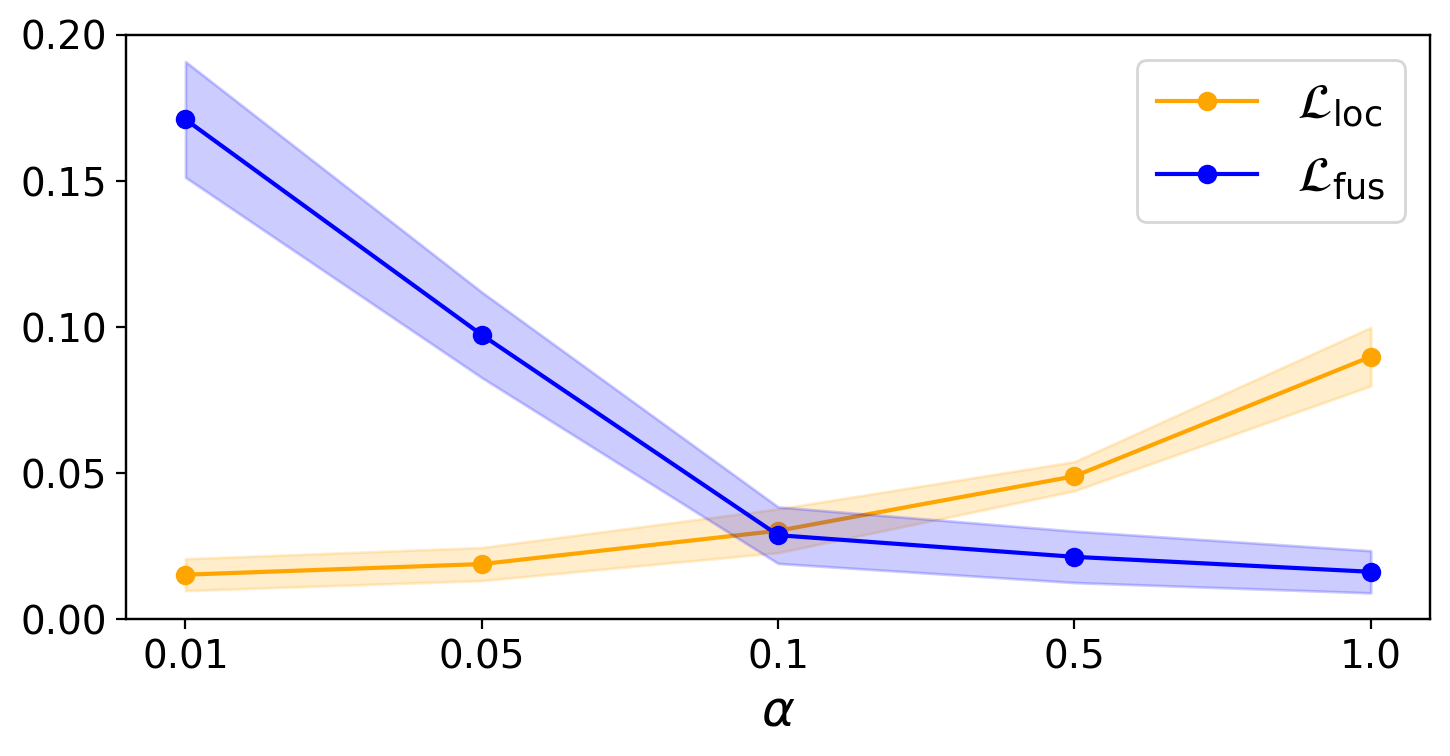}
\caption{The impact on the convergence of $\mathcal{L}_{\rm loc}$ and $\mathcal{L}_{\rm fus}$ by varying $\alpha$. The points represent the mean values of the two losses in the last 1,000 iterations, while the width of the shaded area indicates the standard variance.}
\label{loss_alpha}
}
\end{figure}

\subsection{Hyperparameters} 
\noindent \textbf{$d$ in CARP.}
In CARP, the parameter $d$ determines the number of patches to focus on for each category (real or fake).
As it controls the core pooling operation in the proposed CARP module, we present ablation by varying its value to investigate the impact on performance. 
As shown in Fig. \ref{carp_d}, increasing $d$ improves the performance of the evaluation model, which benefits from richer forgery cues located by FoCus.
Additionally, the performance gain is less significant when the $d$ is larger than 32. 
Thus, we set $d$ to 32.

\noindent \textbf{$\alpha$ in the object function.}
In Equation \ref{total_object}, the parameter $\alpha$ determines the weights of $\mathcal{L}_{\rm fus}$.
We set the value of $\alpha$ based on the convergence of the two losses to make them converge to the same order of magnitude. 
In our implementation, we set $\alpha$ as 0.1. 
We provide the convergence of the two losses with the change of $\alpha$ in the Fig. \ref{loss_alpha}.
Increasing $\alpha$ to 1 or decreasing $\alpha$ to 0.01 harms the convergence of either $\mathcal{L}_{\rm loc}$ or $\mathcal{L}_{\rm fus}$. 
Therefore, setting $\alpha$ as 0.1 helps to approach a Pareto equilibrium between these two losses.

\subsection{Inference performance} 
\begin{table}[t]
\centering
\caption{In-dataset evaluation on FF++(HQ) and cross-dataset evaluation on CDF, DFD, DFDC, DFDCP, and WDF. All models are trained using FF++(HQ). Bold refers to the top performance.
If the results in the original paper are video-level results, we then use their official weights to obtain the frame-level results and indicate it with \dag.}
\label{inference}
\resizebox*{\linewidth}{!}{
\begin{tabular}{ccccccc}
\hline
\multirow{3}{*}{Model}  & \multicolumn{6}{c}{Testing Dataset}                                                 \\ \cline{2-7} 
                                     & \multicolumn{2}{c}{FF++(HQ)}                        & CDF   & DFDC  & WDF   & DFDCp \\ 
                                        & Acc                      & AUC                      & AUC   & AUC   & AUC   & AUC   \\ \hline
MADD\cite{zhao2021multi}                                & 96.70 & 99.29 & 67.44 & 62.70\dag  & 68.03\dag     & 66.28 \\
GFFD\cite{luo2021generalizing}                                & 93.64\dag & 95.06\dag & 75.70\dag & 66.31\dag  & 74.98\dag     & 73.69\dag  \\
DCL\cite{dcl}                               & - & 99.30 & 82.30 & -  & -     & 76.71 \\
UIA-ViT\cite{uiavit}                               & 96.05\dag & 97.98\dag & 82.41 & 72.93\dag  & \textbf{77.98}\dag     & 75.80 \\
SFDG\cite{sfdg}                                    & \textbf{98.19}                    & \textbf{99.53}                    & 75.83 & \textbf{73.64} & 69.27 & -     \\\hline
FoCus(Ours)                                    & 97.28                    & 99.15                    & \textbf{83.17} & 72.98 & 76.02 & \textbf{78.01} \\ \hline
\end{tabular}}
\end{table}

To further analyze our framework, we evaluate the proposed FoCus in terms of the performance of face forgery detection. 
Instead of evaluating different methods through the evaluation model in Fig. \ref{fig4}, we directly test their inference performance for binary classification of face forgeries.
Since there are three classification heads in FoCus, \ie, $\mathbf{\hat{y}}_{\rm fus}$, $\mathbf{\hat{y}}_{\rm RGB}$, and $\mathbf{\hat{y}}_{\rm Sobel}$, we provide the results using $\mathbf{\hat{y}}_{\rm fus}$ for prediction which is based on cross-modal features in Table \ref{inference}.
We note that FoCus demonstrates comparable performance to recent SOTA methods. 
FoCus achieves balanced performance on both in-dataset and cross-dataset evaluation.
For instance, while MADD, DCL, and SFDG perform better than FoCus within the dataset, their cross-dataset performance falls short of FoCus. 
Besides, FoCus achieves the top performance on CDF, DFDC, and DFDCp. 
This validates that locating forgery traces in the fake faces as comprehensively as possible can enhance the model's generalization.
This is consistent with the result in Table \ref{tab2} that the evaluation models supervised by our manipulation maps are more general to unseen face forgeries.

\subsection{computational complexity} 
We provide the computational complexity of our method in detail including FLOPs, throughput on NVIDIA V100, and model parameters in Table \ref{fzd}.
We also list the computational complexity of the three previously compared models, \ie, MADD, GFFD, and UIA-ViT.
Compared to MADD and GFFD, FoCus adopts a smaller input size, making it more efficient during inference. 
Compared to UIA-ViT, which uses ViT Base\cite{dosovitskiy2020image} as its backbone network, FoCus uses two lightweight ViT Small variants in \cite{deit} as the backbone.
Thus, FoCus has a lower parameter amount.
Moreover, FoCus does not contain the hard-to-parallelize matrix inversion operation as UIA-ViT does, resulting in a higher throughput speed.

\begin{table}[t]
\centering
\caption{Computational complexity of our and three compared methods.}
\label{fzd}
\begin{tabular}{ccccc}
\hline
Model &Input size & \#Param.$\downarrow$ & FLOPs$\downarrow$ & \begin{tabular}[c]{@{}l@{}}Throughput\\ (image / s) $\uparrow$\end{tabular}\\ \hline
MADD\cite{zhao2021multi}& $380\times380$ & 49.4M    & \textbf{8.4G} & 220.4 \\
GFFD\cite{luo2021generalizing}& $256\times256$ & 53.3M    & 13.8G & 206.7 \\
UIA-ViT\cite{uiavit}& $224\times224$ & 87.3M    & 16.9G & 242.5 \\
FoCus(Ours)  & $224\times224$  & \textbf{47.4M}    & 8.6G  & \textbf{311.2}\\\hline
\end{tabular}
\end{table}

\section{Conclusion and Future Work}

In this paper, we demonstrate the feasibility of generating exploitable manipulation maps without using paired faces.
Using merely class labels, the proposed FoCus can locate exploitable forged cues to supervise existing multi-task face forgery detection models.
The proposed FoCus benefits from utilizing two image modalities to generate manipulation maps. 
The seed regions in both modalities are located in the proposed CARP module specifically tailored for discovering forgery cues, and then fused using the carefully designed Complementary Learning module.
Visualization results demonstrate the interpretability and robustness of maps generated by our FoCus and the exploitability of the located forgery cues.
Extensive experiments verify the effectiveness of the proposed FoCus in providing pixel-level annotation for auxiliary tasks in face forgery detection.

Recent pixel-level face forgery detection works have also begun to rethink the effectiveness of the manipulation maps obtained by comparison\cite{uiavit,miao2023multispectral}.
Compared to these works, the novelty of our method lies in that it explicitly generate manipulation maps and evaluate them through training multi-task models.
However, we remain aware of the limitation that the manipulation maps produced by FoCus may not be the optimal manipulation maps to serve as supervision.
Nevertheless, none of the current existing face forgery datasets offer ground truth to evaluate the generated manipulation maps.
Thus, the contribution of the proposed FoCus remains significant as it offers better supervision to improve multi-task face forgery detectors as shown in quantitative experiments.

Under the circumstance that pixel-level face forgery detection requires better manipulation maps, the protocol for evaluating manipulation maps is needed.
We conducted evaluations by using the produced manipulation maps as supervision on four existing face forgery detectors and the framework in Fig. \ref{fig4}.
In the future, researchers can consider designing protocols to evaluate manipulation maps from perspectives beyond serving as supervision in segmentation training.
Considering that pixel-level face forgery detection models are often used to provide interpretable detection results, we can emulate works in large language models to construct a human-preference model for assessing the interpretability of manipulation maps.

\newpage

 




\vfill

\end{document}